\documentclass[letterpaper]{article} 
\usepackage{aaai23}  
\usepackage{times}  
\usepackage{helvet}  
\usepackage{courier}  
\usepackage[hyphens]{url}  
\usepackage{graphicx} 
\usepackage{natbib}  
\usepackage{caption} 
  \newcommand{\sign}[1]{\mathrm{sgn}(#1)}
\usepackage{algorithm}
\usepackage{algorithmic}
\usepackage{amsfonts,amssymb}
\usepackage{amsmath, xparse}
\usepackage{booktabs}
\usepackage{multirow}
\usepackage[colorlinks,
            linkcolor=red,      
            anchorcolor=black,  
            citecolor=black,      
            ]{hyperref}
\usepackage{breakurl}
\usepackage{newfloat}
\usepackage{listings}
\DeclareCaptionStyle{ruled}{labelfont=normalfont,labelsep=colon,strut=off} 
\floatstyle{ruled}
\newfloat{listing}{tb}{lst}{}
\floatname{listing}{Listing}

\title{Polarimetric Inverse Rendering for Transparent Shapes Reconstruction}
\author{
    Mingqi Shao,
    Chongkun Xia,
    Dongxu Duan,
    Xueqian Wang
}
\affiliations{
    Tsinghua Shenzhen Internation Graguate School\\
    
    \{smq21,ddx21\}@mails.tsinghua.edu.cn\\
    \{xiachongkun,wang.xq\}@sz.tsinghua.edu.cn
}

\usepackage{bibentry}
\nocopyright
\begin{document}

\maketitle

\begin{abstract}
In this work, we propose a novel method for the detailed reconstruction of transparent objects by exploiting polarimetric cues. Most of the existing methods usually lack sufficient constraints and suffer from the over-smooth problem. Hence, we introduce polarization information as a complementary cue. We implicitly represent the object’s geometry as a neural network, while the polarization render is capable of rendering the object’s polarization images from the given shape and illumination configuration. Direct comparison of the rendered polarization images to the real-world captured images will have additional errors due to the transmission in the transparent object. To address this issue, the concept of reflection percentage which represents the proportion of the reflection component is introduced. The reflection percentage is calculated by a ray tracer and then used for weighting the polarization loss. We build a polarization dataset for multi-view transparent shapes reconstruction to verify our method. The experimental results show that our method is capable of recovering detailed shapes and improving the reconstruction quality of transparent objects.  Our dataset and code will be publicly available at \href{https://github.com/shaomq2187/TransPIR}{https://github.com/shaomq2187/TransPIR}.
\end{abstract}

\section{Introduction}
The acquisition of transparent 3D shapes has always been a challenge in computer vision since their visual appearance is determined by the light paths with both reflected and refracted light. The complex optical characteristic of transparent shapes makes the observation of the RGB camera or the commonly used depth camera contain significant error and leads to the poor performance of the traditional 3D reconstruction methods such as active camera scanning and multi-view stereo.

Recently, the neural inverse rendering based on the neural implicit representation\cite{michalkiewicz2019implicit,oechsle2021unisurf,yariv2020multiview,yariv2021volume} has achieved charming performance in the task of learning 3D shapes from 2D images of opaque objects. However, the high complexity of transparent objects' light paths couples the color of the transparent surface with its geometry, environment light, and viewing direction, leading to the difficulty of applying the neural inverse rendering methods that exploit the photometric consistency constraint, e.g., IDR\cite{yariv2020multiview}, to transparent shapes. In addition, the reconstructed shapes that only use the silhouette constraint suffer from the over-smooth problem. Li et al.\cite{li2020through} proposed a physical-based neural network to handle the complex optical characteristic of transparent shapes under natural lighting conditions, however, it still suffers from the same over-smooth problem. Therefore, additional cues are necessary to be introduced to recover the detailed shapes.
\begin{figure}[t]
\centering 
\includegraphics[width=\linewidth]{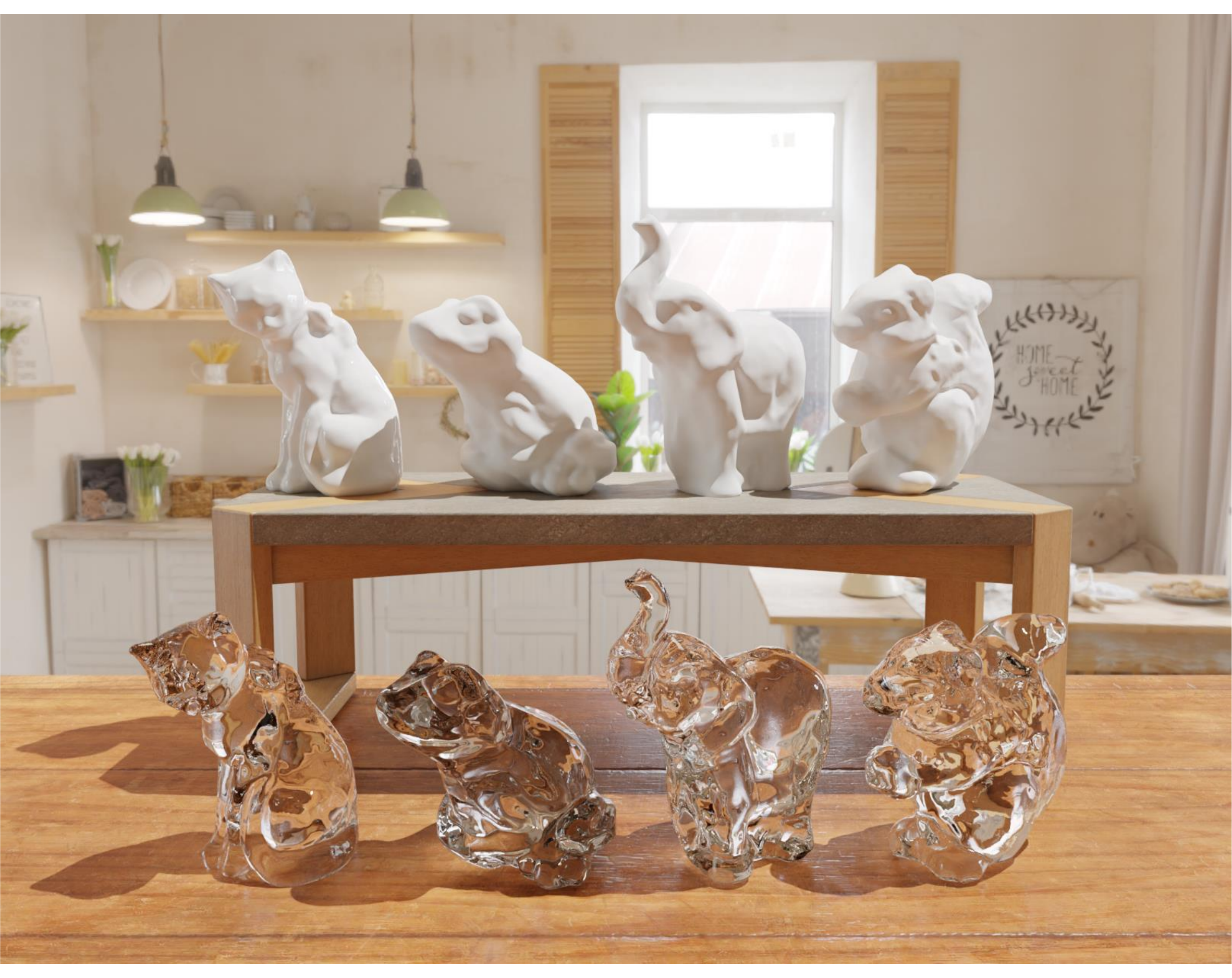} 
\caption{\textbf{Our reconstruction results and their transparent renderings}} 
\label{fig:introduction} 
\end{figure}
Researchers introduce the ray-ray correspondence between camera rays and the rays from the background pattern as the light path's constraint as a cue for detailed shape reconstruction\cite{kutulakos2008theory,tsai2015does,qian20163d,wu2018full,lyu2020differentiable}. However, the ray-ray correspondence generally requires strict calibration and precise control. 

As a passive imaging principle with weak assumptions of lighting conditions, polarimetric cues perform well on many tasks\cite{ba2020deep,zhao2020polarimetric,deschaintre2021deep} since the polarimetric cues provide light's information from a new dimension in addition to intensity. The polarization state of the reflected light from the object's surface encodes the azimuth and zenith angle of the surface normal. Therefore, this paper introduces polarization information as an additional cue for transparent shapes reconstruction. 

Due to the complex light interactions of transparent objects, the useful polarization state in the directly reflected light, encoding the normal vector of the surface, is frequently disturbed by the transmitted light. To reduce the negative contribution of severely disturbed areas' polarization state to the reconstructed shape, we employ ray tracing to trace the proportion of reflected light at each point and apply it as the weight of the polarimetric cue.

In this paper, the neural implicit representation is used to represent the object's geometry. The polarization maps at each view are rendered through a differentiable polarimetric renderer, and the reflection percentage is calculated through a ray tracer, which is used as the weight of the polarization loss. Fig.\ref{fig:introduction} shows the reconstruction results of our method. Experimental results illustrate that our method can reconstruct detailed transparent shapes. We summarize our contributions as follows:
\begin{itemize}
\item A polarimetric inverse rendering framework for transparent shape reconstruction from multi-view polarization images.
\item A weighted polarization loss which utilizes the reflection percentage enables the polarimetric cues can be effectively used.
\item Producing the detailed reconstruction results of different transparent objects with complex and irregular shapes.  
\item The first polarization dataset for multi-view transparent shapes reconstruction.
\end{itemize}

\section{Related Work}
\subsection{Transparent Shapes Reconstruction}
The main challenge of transparent shape reconstruction is that their surface observation will be interfered by the transmission from the background. This feature is exploited by researchers to place known patterns on the back of transparent objects to obtain the correspondence between the emitted ras from transparent objects and the ras from the background patterns\cite{kutulakos2008theory,wu2018full,lyu2020differentiable}. In contrast to the methods that require strict control of the settings to get accurate rays correspondence, our method only needs a dark background and uniform light intensity in directions facing the transparent object.


Recently, data-driven approaches show advantages for shape reconstruction of transparent objects. Li et al.\cite{li2020through} propose a neural 3d reconstruction framework for transparent objects, which simulates light transport within transparent objects through a physically-based rendering layer and uses a pre-trained network for point cloud reconstruction. However, the gap between synthetic dataset and real-world data and insufficient constraint of RGB information lead to the over-smooth phenomenon of the reconstructed shapes\cite{xu2022hybrid}. Additional information is needed to enhance the constraint.

The polarization information is an important cue for transparent shapes estimating\cite{miyazaki2004transparent,mingqi2022transparent} because the polarization state of the light reflected from an object's surface encodes the information of the surface normal\cite{atkinson2006recovery}. However, the methods purely rely on the polarization information usually suffer from large errors since the reflected light, which encodes the surface normal vector, is generally covered by other light components. Hence, different from these methods purely relying on polarization information, our method serves polarization information as cues for multi-view reconstruction to provide auxiliary constraints.
\begin{figure*}[tbp]
\centering 
\includegraphics[width=\textwidth]{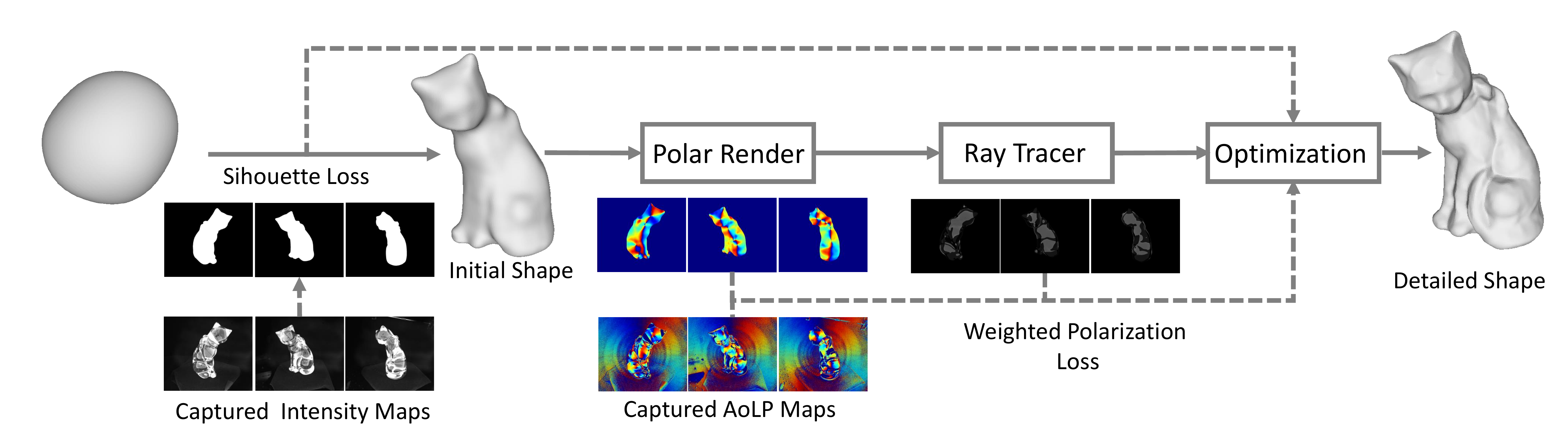} 
\caption{\textbf{Overview of our method.} The silhouette loss is used as supervision of initial shape reconstructing, then the polarimetric render and ray tracer calculate the weighted polarization loss for detailed shape optimization} %
\label{fig:overview} 
\end{figure*}

\subsection{Polarimetric Inverse Rendering}
The polarization state of an object surface encodes the information of the surface normal and is often used for surface estimation\cite{miyazaki2005inverse,zou20203d,ba2020deep},reflection removal\cite{lei2020polarized}, radiance decomposition\cite{deschaintre2021deep,dave2022pandora}, multi-view stereo enhancement\cite{cui2017polarimetric,yang2018polarimetric,ding2021polarimetric}. Miyazaki et al.\cite{miyazaki2005inverse} propose inverse polarization ray-tracing for transparent shape reconstruction, but the assumption that the background shape is known limits its application. Zhao et al.\cite{zhao2020polarimetric} propose a polarimetric multi-view inverse rendering framework for 3D reconstruction, this framework uses polarization information to optimize each vertex of the initial model generated from structure-from-motion to reconstruct detailed shape. Our key idea is similar to \cite{zhao2020polarimetric}, but the polarization information availability of each point on the transparent surface needs to be judged.    

\section{Method}
\subsection{Overview}
Our goal is to recover the transparent shape from multi-view 2D images by exploiting polarimetric cues and the pipeline of our method is shown in Fig.\ref{fig:overview}. Instead of starting with the space carving method\cite{kutulakos2000theory}, we adopt the neural implicit representation in IDR\cite{yariv2020multiview} to produce a smooth initial shape. The polarimetric render will render the AoLP maps of different views and then the rendered AoLP maps will be compared with the captured AoLP maps to get polarization loss. Since only the polarization information of points with a high proportion of reflection component is reliable, the ray tracer will trace the reflection percentage of each point to weight the polarization loss. Afterward, the weighted polarization loss guides the optimization to produce the final detailed shape.

\subsection{Implicit Surface Representation}
Signed distance function(SDF) is a continuous function of spatial position, for a given position $x\in\mathbb{R}^3$, the SDF will output the closest distance $d\in\mathbb{R}$ to the surface. 
\begin{equation}  
	f: \mathbb{R}^3\rightarrow\mathbb{R}  \qquad   x\rightarrow d= f(x)
\end{equation} 
The distance $d$ is positive when $x$ is inside the boundary, negative outside, and zero when $x$ is on the boundary. 

Similar to IDR\cite{yariv2020multiview}, we represent the transparent objects' geometry as a neural network(MLP) $f_\theta$ with learnable parameters $\theta$ and optimize $f_\theta(x)$ to the object's ground-truth SDF $f(x)$:
\begin{equation}  
	opt: f_\theta(x) \rightarrow f(x)
\end{equation}
The normal $\hat{n}_\theta(x)$ of the surface represented by MLP-based SDF $f_\theta(x)$ can be expressed as follows:
\begin{equation}  
	\hat{n}_\theta(x) = \nabla_xf_\theta(x)/\Vert\nabla_xf_\theta(x)\Vert_2
\end{equation} 
The derivative of $f_\theta(x)$ can be easily acquired from the automatic differentiation mechanism.

The SDF representation has the benefits of being able to represent smooth surfaces and arbitrary topologies. With the SDF representation and the supervision of silhouettes, the transparent object's initial shape can be obtained.
\subsection{Polarimetric Rendering}
In this paper, we use the Mueller calculus\cite{clarke2009stellar} to calculate the polarization state of the specular reflection of the transparent surface. In the Muller calculus, the full polarization state of light is represented by the Stokes vector $\mathbf{s}=[s_0,s_1,s_2,s_3]^{\rm T}$, where $s_0$ represents the light intensity, $s_1$ and $s_2$ denote the linear polarization components of the $x$-axis and $45^\circ$ directions, and $s_3$ represents the right circular polarization component. The three polarimetric cues intensity $I$, degree of linear polarization(DoLP) $\rho$, angle of linear polarization(AoLP) $\psi$ can be parameterized by the stokes vector:
\begin{equation}  
	I = s_0
\end{equation} 
\begin{equation}  
	\rho = \frac{\sqrt{s_1^2+s_2^2}}{s_0}
\end{equation} 
\begin{equation}  
	\psi = \frac{1}{2}\tan^{-1}(\frac{s_2}{s_1})
	\label{eq:psi}
\end{equation}

\begin{figure}[htbp]
\centering 
\includegraphics[width=\linewidth]{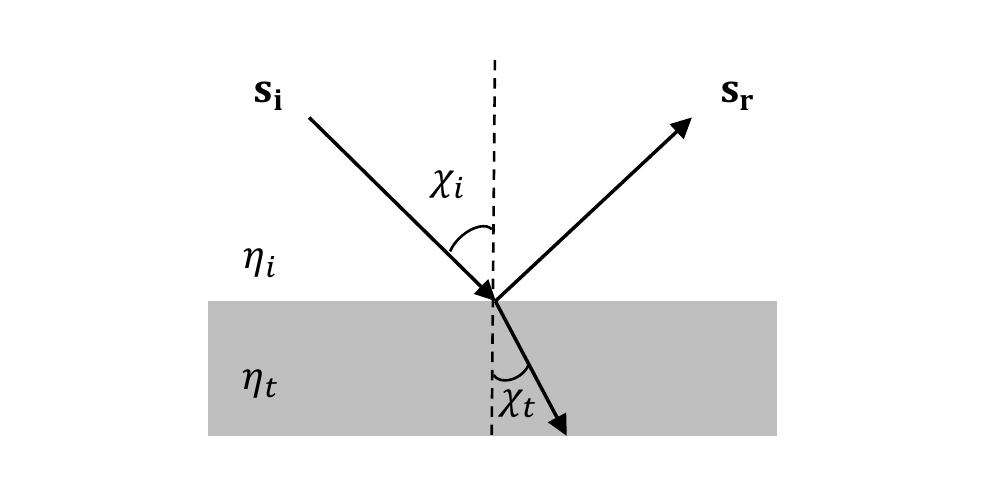} 
\caption{\textbf{Reflection and transmission at the interface of different refractive index media}} 
\label{fig:reflection} %
\end{figure}

From the Fresnel's equations, the amplitude reflection coefficients $r_s, r_p$ that are perpendicular to and parallel to the incident plane, respectively, can be written as follows\cite{clarke2009stellar}:
\begin{equation}  
	r_s = \frac{\eta_i\cos\chi_i-\eta_t\cos\chi_t}{\eta_i\cos\chi_i+\eta_t\cos\chi_t}=-\frac{\sin(\chi_i-\chi_t)}{\sin(\chi_i+\chi_t)}
\end{equation} 
\begin{equation}  
	r_p = \frac{\eta_i\cos\chi_t-\eta_t\cos\chi_i}{\eta_t\cos\chi_i+\eta_i\cos\chi_t}=-\frac{\tan(\chi_i-\chi_t)}{\tan(\chi_i+\chi_t)}
\end{equation} 

As shown in Fig.\ref{fig:reflection}, when the stokes vectors $\mathbf{s_i}, \mathbf{s_r}$ of the incident light and the reflected light are in the same coordinate system, the transformation between $\mathbf{s_i}, \mathbf{s_r}$ can be represented by a mueller matrix $M_r$, that is,
\begin{equation}  
	\mathbf{s_r} = M_r \mathbf{s_i}
\end{equation} 

\begin{equation}
	M_r= \frac{1}{2}
	\begin{bmatrix}
		(r_s^2 + r_p^2) & (r_s^2 - r_p^2) & 0 & 0   \\
		(r_s^2 - r_p^2) & (r_s^2 + r_p^2) & 0 & 0   \\
		0 & 0 & 2r_sr_p & 0   \\
		0 & 0 & 0 & 2r_sr_p   \\		
	\end{bmatrix}
\end{equation}
\begin{figure}[htbp]
\centering 
\includegraphics[width=\linewidth]{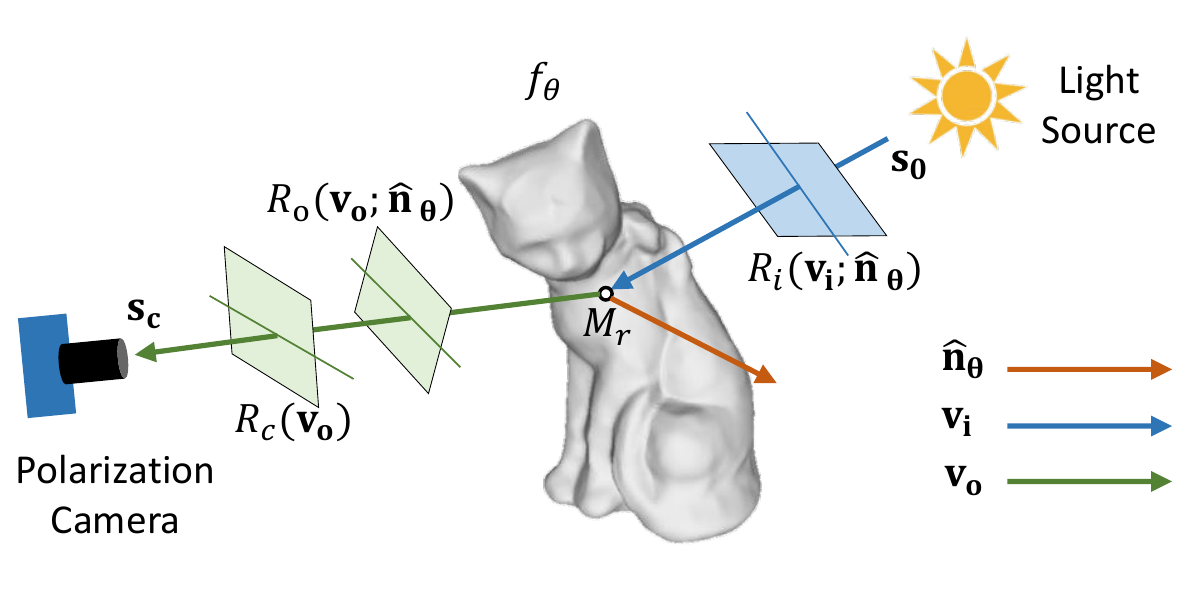} 
\caption{\textbf{Schematic of polarimetric rendering in our method.} The initial Stokes vector from the light source is observed by the camera after four transformations: three coordinate transformations $R_i,R_o,R_c$, and one polarization state transformation $M_r$ of the reflection} 
\label{fig:polar_render} 
\end{figure}

Fig.\ref{fig:polar_render} shows the schematic of polarimetric rendering used in our method. The coordinate frames transformations are required since the stokes vectors $\mathbf{s_0}$ and $\mathbf{s_c}$ are defined in the different reference frames. We adopts the transformations of frames similar to Mitsuba2\cite{nimier2019mitsuba}, and detailed expressions of  $R_i(\mathbf{v_i};\mathbf{\hat{n}_\theta})$, $R_o(\mathbf{v_o};\mathbf{\hat{n}_\theta})$, $R_c(\mathbf{v_o})$ are presented in supplementary material. Finally, the stokes vector $\mathbf{s_c}$ captured by the polarization camera can be written as the linear operations:
\begin{equation}  
	\mathbf{s_c} = R_cR_oM_rR_i \mathbf{s_0} = [s_{c0},s_{c1},s_{c2},s_{c3}]^{\rm T}
	\label{eq:sc}
\end{equation} 

From the Eq.\ref{eq:psi}, the rendered AoLP $\hat{\psi}$ is,
\begin{equation}  
	\hat{\psi} = \frac{1}{2}\tan^{-1}(\frac{s_{c2}}{s_{c1}})
	\label{eq:hat_psi}
\end{equation}

The polar render in Fig.\ref{fig:overview} renders the AoLP maps through Eq.\ref{eq:sc}, Eq.\ref{eq:hat_psi}, then they will be compared with the real-world captured AoLP maps to calculate polarization loss, which will be described in detail in later subsection \textbf{Optimization}. 

\begin{figure}[tbp]
\centering 
\includegraphics[width=\linewidth]{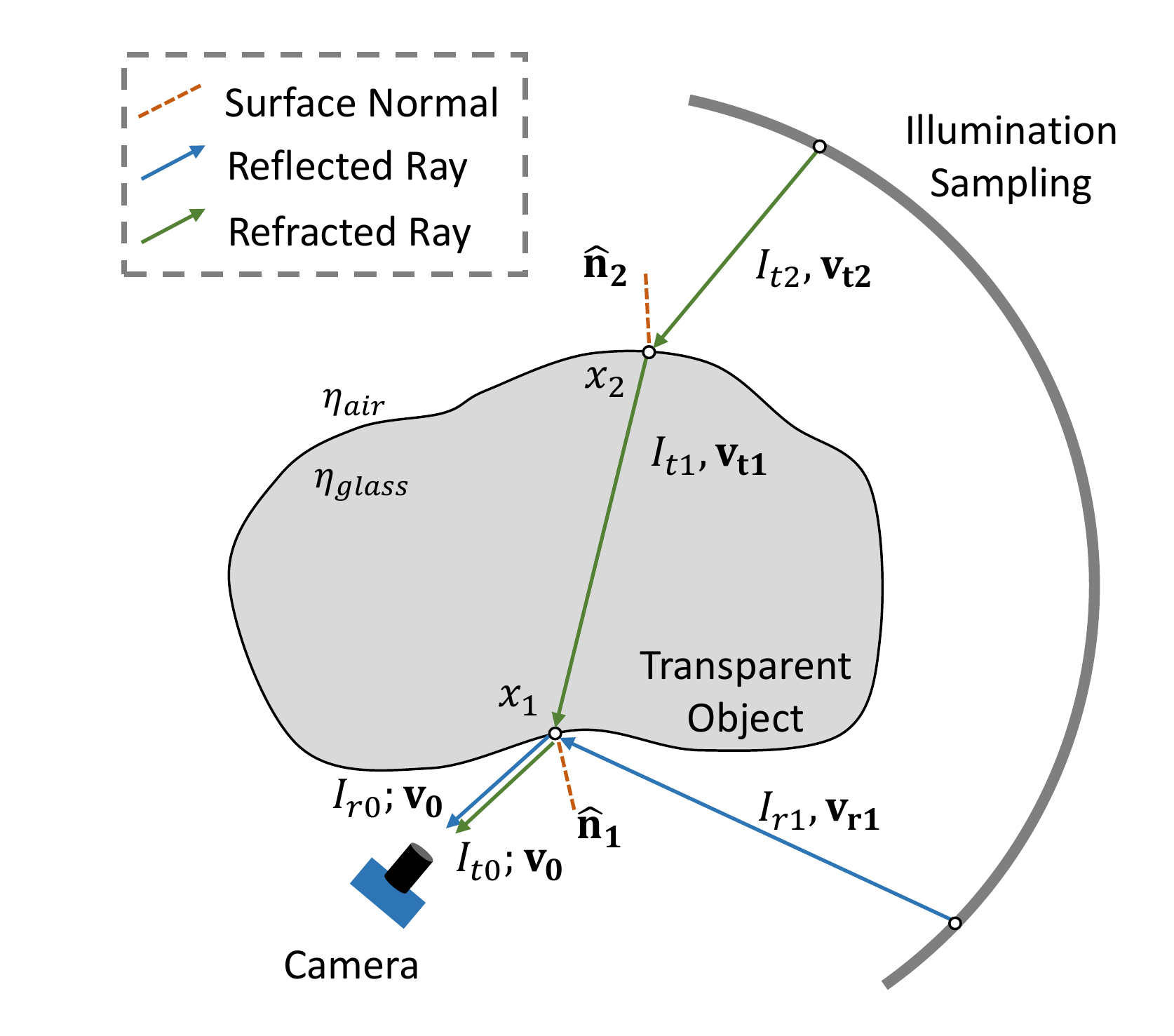} 
\caption{\textbf{Ray tracing procedure.} For the given shape, the illumination configuration, and the viewing direction $-v_0$, the radiance of reflected and transmitted rays $I_{r0}, I_{t0}$ along the $v_0$ direction can be calculated from Fresnel's laws} 
\label{fig:ray_tracer} 
\end{figure}

\subsection{Reflection Percentage Estimation}
\begin{figure*}[tbp]
\centering 
\includegraphics[width=\linewidth]{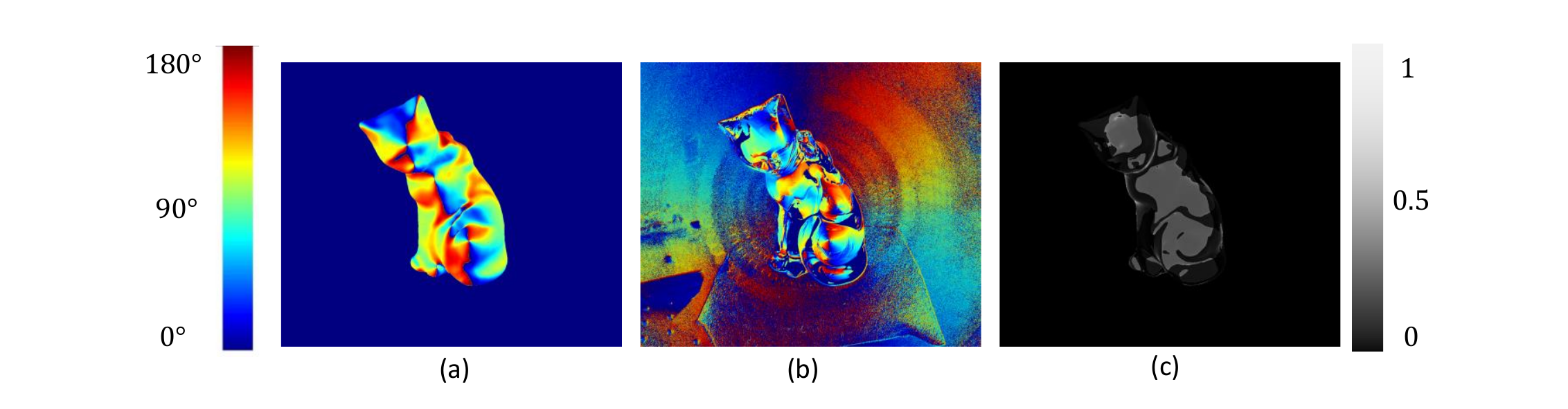} 
\caption{\textbf{(a) The AoLP map rendered by the polar render. (b)Real-world captured AoLP map. (c)The reflection percentage map rendered by the ray tracer.} The area with a low reflection percentage in rendered AoLP map suffers a large error compared to the real-world captured AoLP map}
\label{fig:percentage_importance} 
\end{figure*}

The polar render employs the Fresnel specular reflection model to render the AoLP maps for the given shape. However, in some areas, the rendered AoLP map has a large error compared with the AoLP map captured in the real world as shown in Fig.\ref{fig:percentage_importance}. This error is mainly caused by the high transmission rather than shape difference. The observed light from the transparent surface consists of two components: the directly reflected light on the surface(the specular reflection component) and the transmitted light from the inside(the transmission component). The AoLP of the specular component is only related to the normal of the intersection point on the surface, while the AoLP of the transmission component is related to the normals of all intersection points in its transmission process. Therefore, the higher proportion of the specular reflection, the more reliable the captured AoLP. To quantify and reduce the resulting error, we use ray tracing to calculate the reflection percentage of each point on the transparent surface. 

We implement a 2-bounce ray tracer to estimate the radiance of reflection and transmission components on the transparent surface. Fig.\ref{fig:ray_tracer} illustrates the ray tracing procedure. The rays in this Fig.\ref{fig:ray_tracer} flow from the light source to the camera, which is inverse in our actual implementation. Given the shape, illumination configuration, and the camera viewing direction $\mathbf{-v_0}$,  the directions of other rays can be derived from Fresnel's law of refraction. In each interaction, the energy of the incident light is distributed according to the Fresnel term $\mathcal{F}$\cite{born2013principles},
\begin{equation}  
	\mathcal{F}_{\eta_i,\eta_t}^{\mathbf{v_i},\mathbf{v_t},\mathbf{n}}= \frac{1}{2}(\frac{\eta_i\mathbf{v_i}\cdot\mathbf{n}-\eta_t\mathbf{v_t}\cdot\mathbf{n}}{\eta_i\mathbf{v_i}\cdot\mathbf{n}+\eta_t\mathbf{v_t}\cdot\mathbf{n}})^2 + \frac{1}{2}(\frac{\eta_t\mathbf{v_i}\cdot\mathbf{n}-\eta_i\mathbf{v_t}\cdot\mathbf{n}}{\eta_t\mathbf{v_i}\cdot\mathbf{n}+\eta_i\mathbf{v_t}\cdot\mathbf{n}})^2
	\label{eq:frenel_term}
\end{equation}
\begin{equation}  
	I_r = \mathcal{F}_{\eta_i,\eta_t}^{\mathbf{v_i},\mathbf{v_t},\mathbf{n}} I_i
	\label{eq:frenel_term_It}
\end{equation}
\begin{equation}  
	I_t = (1-\mathcal{F}_{\eta_i,\eta_t}^{\mathbf{v_i},\mathbf{v_t},\mathbf{n}}) I_i
	\label{eq:frenel_term_Ir}
\end{equation}
where $I_i, I_r, I_t$ are the intensity of incident, reflected and refracted  rays, $\mathbf{v_i}, \mathbf{v_r}, \mathbf{v_t}$ represent their directions, respectively. $\mathbf{n}$ is the normal vector.

Using Eq.\ref{eq:frenel_term}, \ref{eq:frenel_term_It}, \ref{eq:frenel_term_Ir}, we can calculate the intensities of reflected ray and transmitted ray $I_{t0}, I_{r0}$ observed by the camera in Fig.\ref{fig:ray_tracer}:
\begin{equation}  
	I_{t1} = (1-\mathcal{F}_{\eta_{air},\eta_{glass}}^{\mathbf{v_{t2}},\mathbf{v_{t1}},\mathbf{\hat{n}_2}})I_{t2}
\end{equation}
\begin{equation}  
	I_{t0} = (1-\mathcal{F}_{\eta_{glass},\eta_{air}}^{\mathbf{v_{t1}},\mathbf{v_{0}},\mathbf{\hat{n}_1}})I_{t1}
\end{equation}
\begin{equation}  
	I_{r0} = \mathcal{F}_{\eta_{air},\eta_{glass}}^{\mathbf{v_{r1}},\mathbf{v_{0}},\mathbf{\hat{n}_1}}I_{r1}
\end{equation}
where the values of $I_{t2}, I_{r1}$ are sampled from the illumination configuration, which is related to our dataset acquisition setup and is presented in the supplementary material in detail. Due to the total internal reflection, the values of $I_{t1}$ of some rays are unable to calculate from the 2-bounce ray tracer, in this case, we set $I_{t1}$ to a constant to avoid 100\% reflection percentage since it is impossible in the real world.

Finally, the reflection percentage $w$ in this paper is defined as the ratio of the reflection intensity to total intensity:
\begin{equation}  
	w = \frac{I_{r0}}{I_{r0}+I_{t0}}
\end{equation}

As shown in Fig.\ref{fig:percentage_importance}, the real-world captured AoLP map (Fig.\ref{fig:percentage_importance}(b)) has obvious demarcation in some areas compared to the rendered AoLP map (Fig.\ref{fig:percentage_importance}(a)) since the proportion of reflection component changes. This change is consistent with the reflection percentage map(Fig.\ref{fig:percentage_importance}(c)) rendered by the ray tracer, which further illustrates the importance of the reflection percentage. Otherwise, the error caused by the smaller reflection proportion will guide the optimization to a wrong shape.
\subsection{Optimization}\label{sec:optimization}
We minimize the following loss function to optimize the MLP-based SDF $f_\theta(x)$ to the object's ground-truth SDF $f(x)$:
\begin{equation}  
	\mathcal{L}_{net} = \mathcal{L}_{sil} + \lambda_{sdf} \mathcal{L}_{sdf} + \lambda_{pol} \mathcal{L}_{pol}
\end{equation}
where $\mathcal{L}_{sil}, \mathcal{L}_{sdf}, \mathcal{L}_{pol}$ represent the silhouette loss, sdf regularization loss term, and weighted polarization loss term. The default values of $\lambda_{sdf}$ and $\lambda_{pol}$ are $0.1$ and $0.4$, respectively.
\subsubsection{Silhouette Loss.} We adopt the same silhouette loss as in IDR\cite{yariv2020multiview} for shape optimization supervision, which plays an important role in initial shape reconstruction.
\subsubsection{SDF Regularization Loss.} To encourage the $f_\theta(x)$ approximates a signed distance function, we add the regularization loss, i.e., the Eikonal regularization\cite{gropp2020implicit}:
\begin{equation}  
	\mathcal{L}_{sdf} = \frac{1}{\Vert P\Vert_1}\sum_{p\in P}(\Vert\nabla f_\theta(p)\Vert_2-1)^2 
\end{equation}
 where $P$ represents the set of intersection points of sampled rays in the mini-batch with the surface. 
\subsubsection{Weighted Polarization Loss.} We use polarimetric cues to calculate polarization loss to guide the optimization. The weighted polarization loss for each sampled ray is defined as follows:
\begin{equation}  
	\mathcal{L}_{pol}^p = w_p\Vert \hat{\psi}_p - \psi_p \Vert_1, ~~~~~p\in P
\end{equation}
where $w_p$ is the reflection percentage of the point $p$. $\hat{\psi}_p$ represents the rendered AoLP, and $\psi_p$ denotes the real-world captured AoLP. As mentioned before, the reflection percentage is related to the reliability of the $\psi_p$, hence we employ it as the weight for the error $\Vert \hat{\psi}_p-\psi_p\Vert_1 $.

We assume that the difference of the normal between the initial shape and the ground-truth shape is smaller than $\varepsilon$ since the supervision of $\mathcal{L}_{sil}$ can produce a good initial shape. With this assumption, we can clip the excessive polarization loss to avoid guiding the optimization to the wrong shape:

\begin{equation}  
	{\mathcal{L}_{pol}^p}' = \begin{cases}
	\mathcal{L}_{pol}^p,& \text{ $ \Vert \hat{\psi}_p - \psi_p \Vert_1 \leq \varepsilon$ } \\
	0,& \text{$otherwise$}
	\end{cases}
\end{equation}
\begin{equation}  
	\mathcal{L}_{pol} = \frac{1}{\Vert P \Vert_1}\sum_{p \in P} {\mathcal{L}_{pol}^{p}}'
\end{equation}
where the default value of $\varepsilon$ in this paper is $\pi/6$.

\section{Experiments}
\subsection{Dataset and Metrices}
\begin{figure}[htbp]
\centering 
\includegraphics[width=0.93\linewidth]{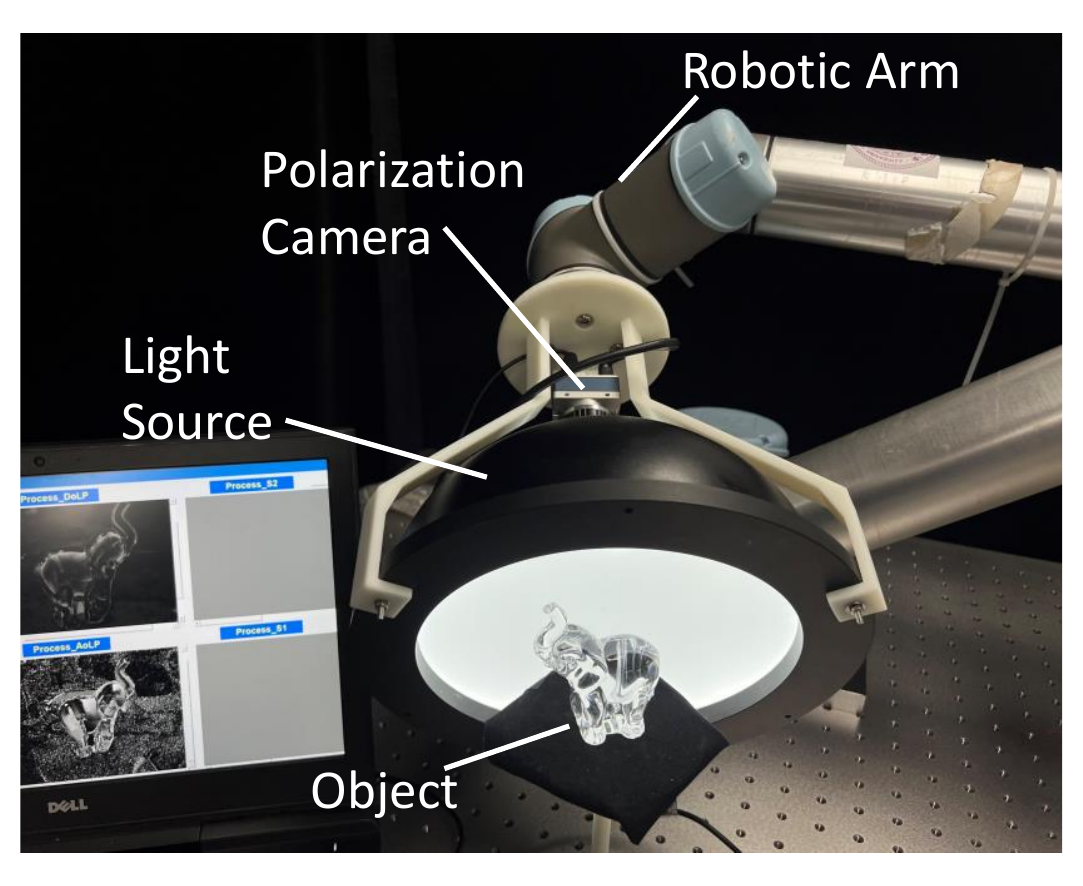} 
\caption{\textbf{Setup for acquiring our dataset}} 
\label{fig:datasetsetup} 
\end{figure}

We build a dataset containing $4$ transparent objects(\textit{CAT, FROG, ELEPHANT, SQUIRREL}) to verify our method since there is no public polarization dataset for multi-view reconstruction of transparent objects. Our dataset acquisition setup is shown in Fig.\ref{fig:datasetsetup}. A DLASA G3-GM14-M2450 polarization camera is adopted as our capture device. For each object, we take $34$ polarization images uniformly distributed views. The pose of the camera is accurately obtained by a commercial robot UR5. To increase the proportions of the reflection components, a diffuse sphere is employed as the light source. The light source is fixed with the camera to ensure uniform and strong reflections in any pose, thereby increasing the abundance of polarization information.

We adopt the chamfer distance(CD) and chamfer normal angle(CDN) as the metrics and the metrics are calculated by uniformly sampling $10000$ points from the ground-truth and reconstructed shape and the metrics that appear in this section are the summation of all the sampling points. For convenience, we normalize the camera's poses to ensure that the reconstructed models are all within a unit sphere and all evaluations are performed on this basis.

\begin{table*}[htbp]
  \centering
  \caption{\textbf{Quantitative results of the comparisons with baselines}}
    \begin{tabular}{ccccccccc}
    \toprule
    \multirow{2}[4]{*}{Method} & \multicolumn{2}{c}{CAT} & \multicolumn{2}{c}{FROG} & \multicolumn{2}{c}{ELEPHANT} & \multicolumn{2}{c}{SQUIRREL} \\
\cmidrule{2-9}          & CD    & CDN   & CD    & CDN   & CD    & CDN   & CD    & CDN \\
    \midrule
    VH    & 18.96 & 3143.56 & 34.06 & 2752.86 & 24.26 & 3810.97 & 14.76 & 3081.47 \\
    IDR   & 9.82  & 978.79 & 18.99 & 1152.16 & 12.52 & 1579.04 & 13.94 & 1691.07 \\
    \textbf{Ours} & \textbf{8.97} & \textbf{744.05} & \textbf{17.22} & \textbf{1088.37} & \textbf{11.31} & \textbf{1510.27} & \textbf{13.35} & \textbf{1476.51} \\
    \bottomrule
    \end{tabular}%
  \label{tab:baselines}%
\end{table*}%

\subsection{Comparisons with Baselines}
We compare our method with two 3D reconstruction methods, IDR\cite{yariv2020multiview} and visual hull(space carving) to verify the effectiveness of our method.

\noindent\textbf{IDR.} IDR is one of the state-of-the-art methods for inverse rendering reconstruction using MLP-based implicit representation. The renderer in IDR is only suitable for opaque objects, its differential renderer will diverge when applied to transparent objects. Hence, we remove the RGB loss term in IDR, only silhouette loss and regularization loss are used. All the other super parameters are the same as in the original IDR.

\noindent\textbf{Visual Hull(VH).} Visual hull or space carving is a traditional algorithm for 3D reconstruction and is usually used as the initial shape reconstruction. We utilize the visual hull code from Li et al.\cite{li2020through} to compute the visual hull. Since our camera poses are already normalized, we limit the visual hull to $[-1,1]^3$ and set the spatial resolution to $256$.

\begin{figure}[htbp]
\centering 
\includegraphics[height=\linewidth]{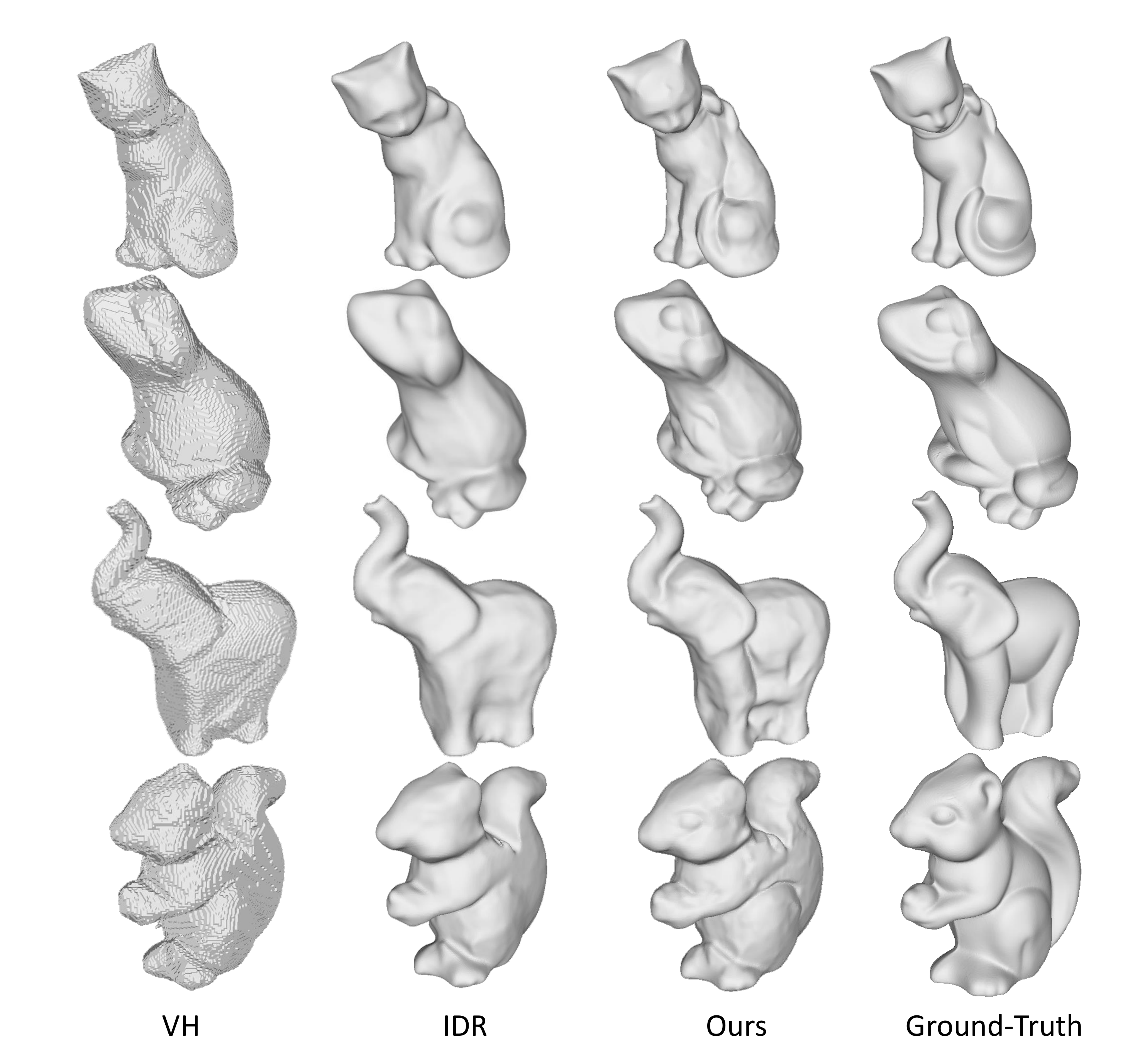} 
\caption{\textbf{Visualization of the reconstruction results compared with baselines}}
\label{fig:baselines} 
\end{figure}

Fig.\ref{fig:baselines} is the visualization of the comparisons and it shows that due to the limitation of the resolution, the visual hull can only reconstruct the rough outline of the object, and its reconstructed surfaces are rough and lack details. In contrast, IDR exhibits the advantage of using MLP-based implicit representation, which can produce water-tight surfaces. However, due to only silhouette supervision, the reconstruction results of IDR also lack details about objects. Our method introduces polarimetric cues as compliments, hence our method has more detailed results, such as the eyes of the objects, compared to IDR. Table.\ref{tab:baselines} lists the quantization results of our comparisons, and our method achieves the best performance on each object.
\subsection{Ablation Studies}
\begin{figure}[htbp]
\centering 
\includegraphics[width=\linewidth]{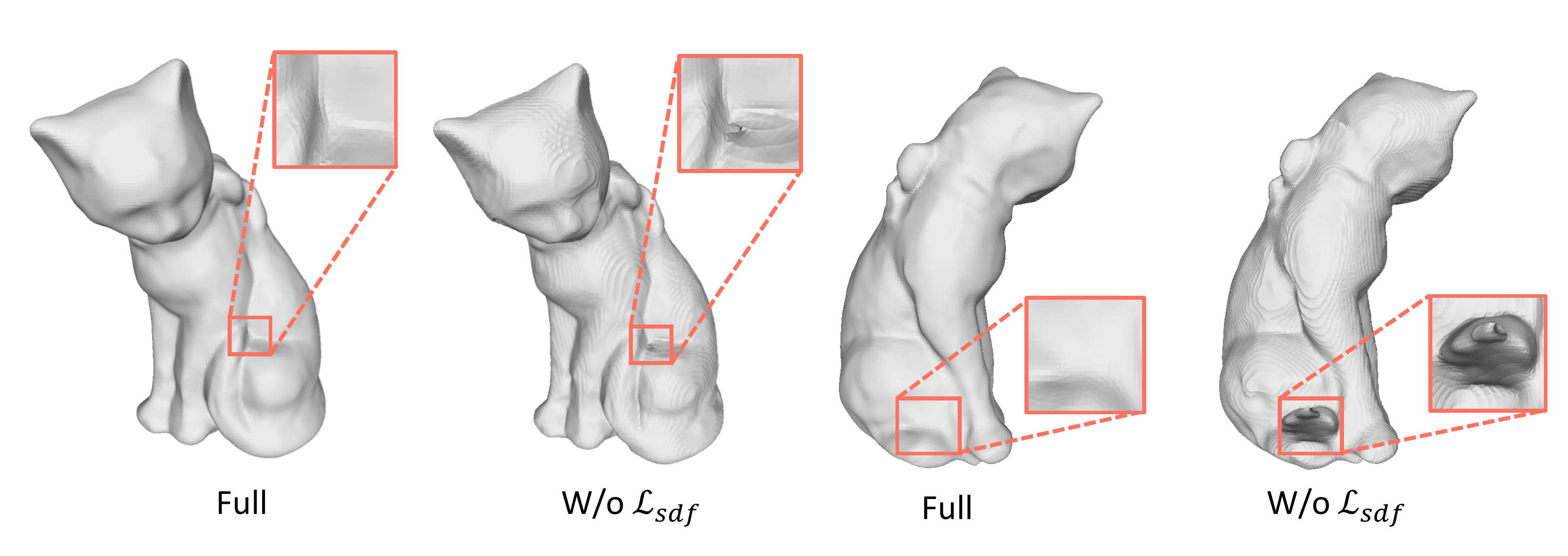} 
\caption{\textbf{Ablation study of the SDF loss term}} 
\label{fig:ablation-study-sdf} 
\end{figure}

We conduct ablation experiments on the important parts of our method, including loss terms $\mathcal{L}_{pol}$, $\mathcal{L}_{sdf}$, and the reflection percentage $w$. In the ablation studies, only the weight of the studied module is set to zero, other parameters are kept the same, and the CAT object is selected for ablation studies. The ablation results of $\mathcal{L}_{pol}$ have been shown in Fig.\ref{fig:baselines} and Table.\ref{tab:baselines} in the previous subsection, i.e., the comparison of ours and IDR, the loss term $L_{pol}$ improves the reconstruction quality by supplementing information of shape's details. 
\begin{table}[htbp]
  \centering
  \caption{\textbf{Quantitative results of the ablation studies}}
  \resizebox{\linewidth}{!}{
    \begin{tabular}{ccccc}
    \toprule
    Metric & W/o $\mathcal{L}_{pol}$ & W/o $\mathcal{L}_{sdf}$& W/o $w$ & Full \\
    \midrule
    CD    & 9.82  & 26.81 & 9.51  & \textbf{8.97} \\
    CDN   & 978.79 & 1788.01 & 1017.79 & \textbf{744.05} \\
    \bottomrule
    \end{tabular}%
    }
  \label{tab:ablation study}%
\end{table}%
Fig.\ref{fig:ablation-study-sdf} shows the results of ablation study on the SDF loss term $\mathcal{L}_{sdf}$. After removing $\mathcal{L}_{sdf}$, obvious contour lines and hollows appear on the reconstructed surface. $\mathcal{L}_{sdf}$ constraints surface normal of the shape that is implicitly represented by a MLP to approach the unit vector, which ensures the reconstructed shape is smooth and realistic. Therefore, when the SDF loss term omits, the MLP will have large gradients at some spatial areas, resulting in holes in the reconstructed shape.

\begin{figure}[htbp]
\centering 
\includegraphics[width=\linewidth]{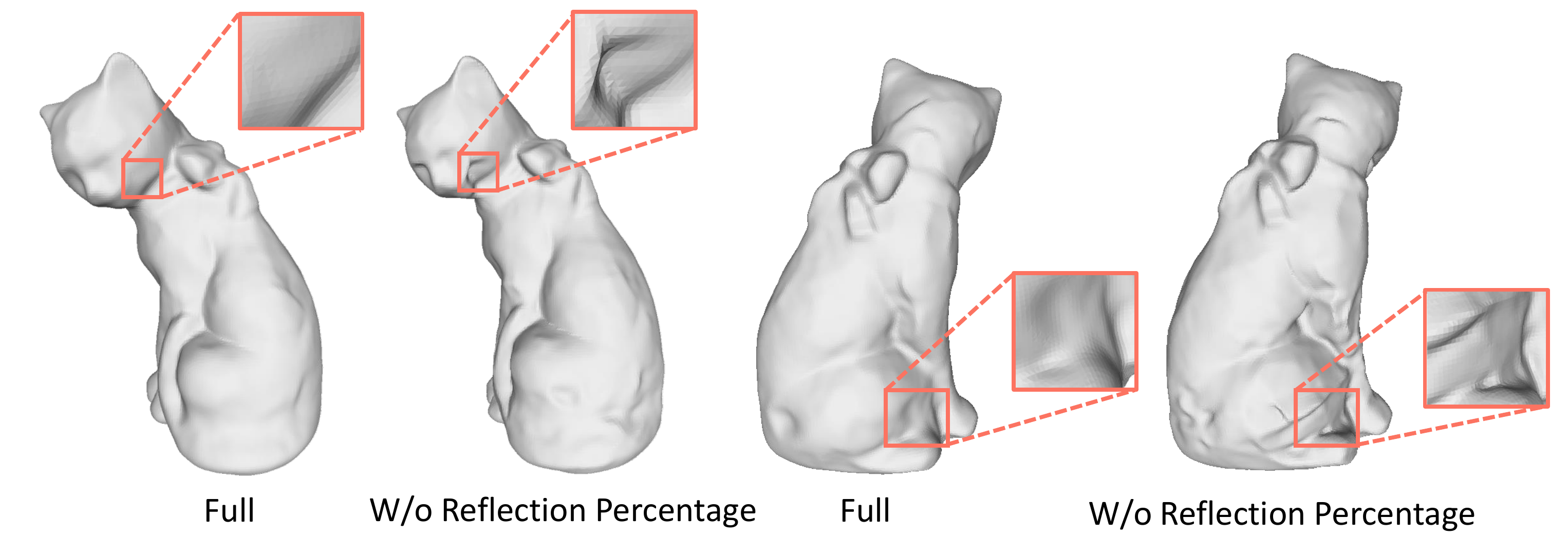} 
\caption{\textbf{Ablation study of the reflection percentage}} 
\label{fig:ablation-study-reflection} 
\end{figure}

Fig.\ref{fig:ablation-study-reflection} presents the results with and without the reflection percentage $w$ and it shows that the polarimetric cues will misguide the shape reconstruction, especially the folded areas. The high transmission component proportion in these areas leads to the coupling of the observed polarization state with all the interaction points in the transmitted light path. Calculating the loss directly with the rendered polarization images will lead to an erroneous shape, which is the reason for introducing the reflection percentage to weight the polarization loss.

Table.\ref{tab:ablation study} lists all the results of our ablation studies. The results illustrate that all the parts in our method are essential to the quality of the reconstructed shape.

\subsubsection{Different Number of Views.}
\begin{figure}[tbp]
\centering 
\includegraphics[width=0.85\linewidth]{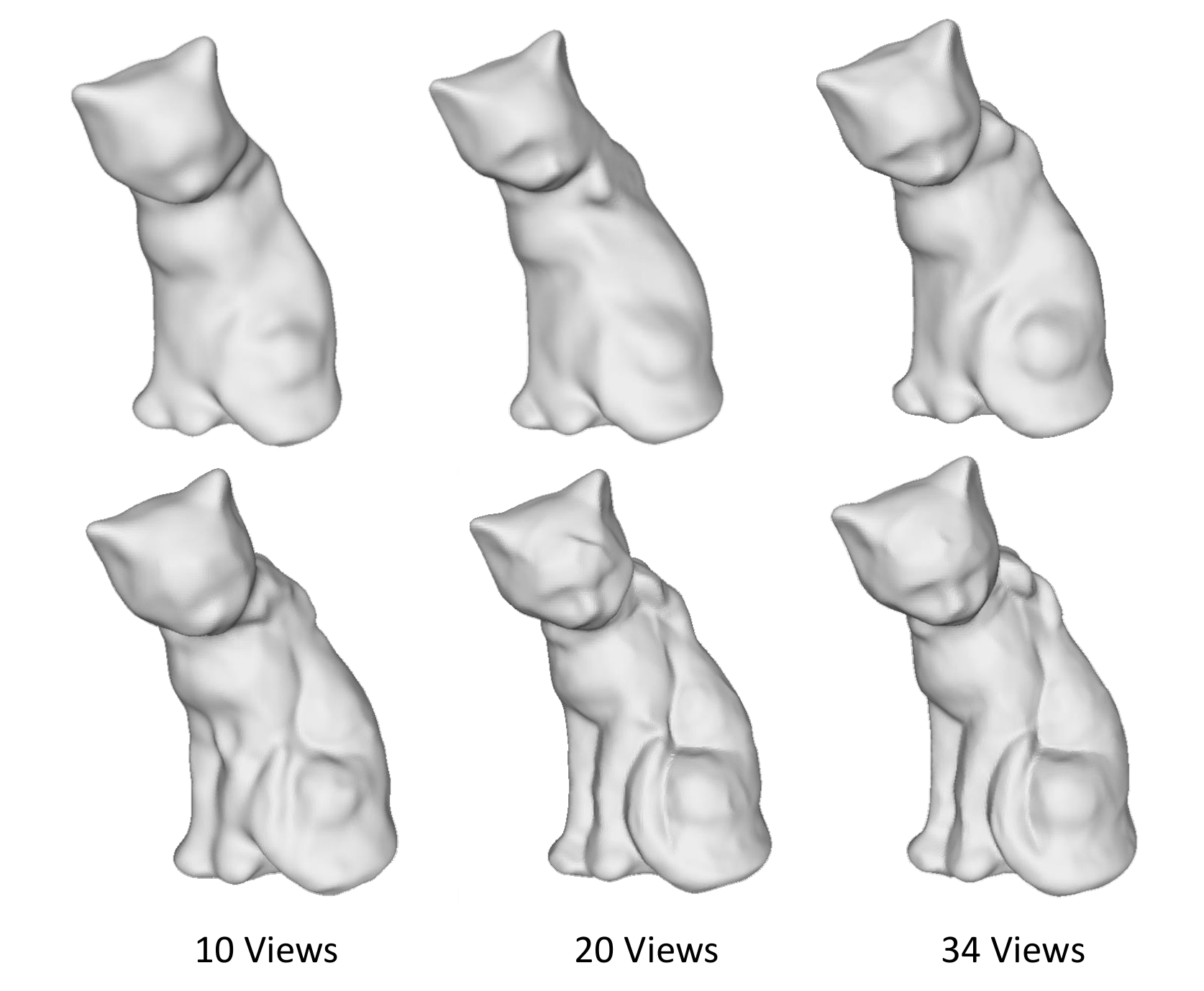} 
\caption{\textbf{Reconstruction results under the different number of views.} The shapes listed in the two rows are the results without/with polarimetric cues, respectively} 
\label{fig:sparseviews} 
\end{figure}

 We uniformly sample $10$ and $20$ views from $34$ views and compare our method with the IDR without polarimetric cues. The visualization and quantitative results of different number of views are summarized in Fig.\ref{fig:sparseviews} and Table.\ref{tab:sparseviews}, respectively. Our method is able to reconstruct the detailed shape under $20$ views but with some noise in the head region, and the quantitative results also show that the difference of chamfer distance between $20$ views and $34$ views is tiny. The details of the reconstructed shape from $10 views$ are significantly reduced due to the reduction of polarimetric cues. But the reconstruction quality of our method outperforms the IDR without polarimetric cues in all the different number of views.
\begin{table}[htbp]
  \centering
  \caption{\textbf{Quantitative results of the reconstructed shapes under the different number of views}}
  \resizebox{\linewidth}{!}{
    \begin{tabular}{ccccccc}
    \toprule
    \multirow{2}[4]{*}{Method} & \multicolumn{2}{c}{10 Views} & \multicolumn{2}{c}{20 Views } & \multicolumn{2}{c}{34 Views} \\
\cmidrule{2-7}          & CD    & CDN   & CD    & CDN   & CD    & CDN \\
    \midrule
    IDR   & 11.75 & 1259.79 & 11.61 & 1371.06 & 9.82  & 978.79 \\
    Ours  & \textbf{9.89} & \textbf{1009.11} & \textbf{9.29} & \textbf{934.67} & \textbf{8.97} & \textbf{744.05} \\
    \bottomrule
    \end{tabular}%
    }
  \label{tab:sparseviews}%
\end{table}%

\section{Discussion}
\noindent \textbf{Conclusion.} In this paper, we propose a polarimetric inverse rendering framework for transparent shapes reconstruction from multi-view polarization images. We employ the implicit neural representation for the object's geometry, then it is rendered by the polarimetric render and compared to the real-world captured polarization images. To address the polarization information reliability reduction caused by the transmission, a ray tracer will trace the reflection percentage to calculate the weighed polarization loss. In addition, we construct the first polarization dataset for multi-view transparent shapes reconstruction and verify our method on this dataset. The experimental results show that our method can recover high-quality transparent shapes, and prove that polarimetric cues can effectively recover the details of transparent objects.

\noindent \textbf{Limitations and future work.} The polarimetric rendering in this paper only considers the reflection component on the transparent surface, i.e., only the polarimetric cues of the areas with high reflection percentage are effectively utilized. Hence, using the polarization ray tracing technique to render more realistic polarization images of transparent objects will be our future work. In addition, the quality of our method heavily depends on the quality of the initial shape, how to reconstruct outperforming shapes based on poor initial shapes is also one of our future directions.
\bibliography{Reference}
\title{111}
\maketitle
\section{Implementation Details} 
We implement the $f_\theta(x)$ as a MLP with $8$ layers and $512$ units per layer in PyTorch\cite{paszke2019pytorch}, and a skip connection from input layer to the middle layer is added as in IDR\cite{yariv2020multiview}. The model is trained on a RTX 3090 GPU(24GB). We use the Adam optimizer\cite{kingma2014adam} with a learning rate of $1e-4$ to optimize the network. We sample $20480$ rays per iteration and employ the secant algorithm to calculate the intersection points of the sampled rays to the shape. Each object is trained for $1000$ epochs, the first $100$ epochs are the initial shape reconstruction stage, and $\lambda_{pol}$ is set to zero. After $100$ epochs,$\lambda_{pol}$ is set to $0.4$($0.2$ for object ELEPHANT) to introduce polarimetric cues. 
\section{Intersection of Ray and Geometry}
In this paper, we employ a neural network to implicitly represent an object's signed distance function(SDF), denoted as $f_\theta(x)$. Both the polarimetric render and ray tracer in our method require the intersections of rays and the geometry, especially in the ray tracer not only requires the intersections of the rays from the camera, but the interactions of the internally refracted rays with geometry also need to be obtained. The ray marching method, which is often used with the SDF representation, can not meet the requirements since it can only be used on one side of the SDF function. Therefore, we adopt the secant algorithm to approach the intersection of rays and geometry.

\begin{figure}[htbp]
\centering 
\includegraphics[width=\linewidth]{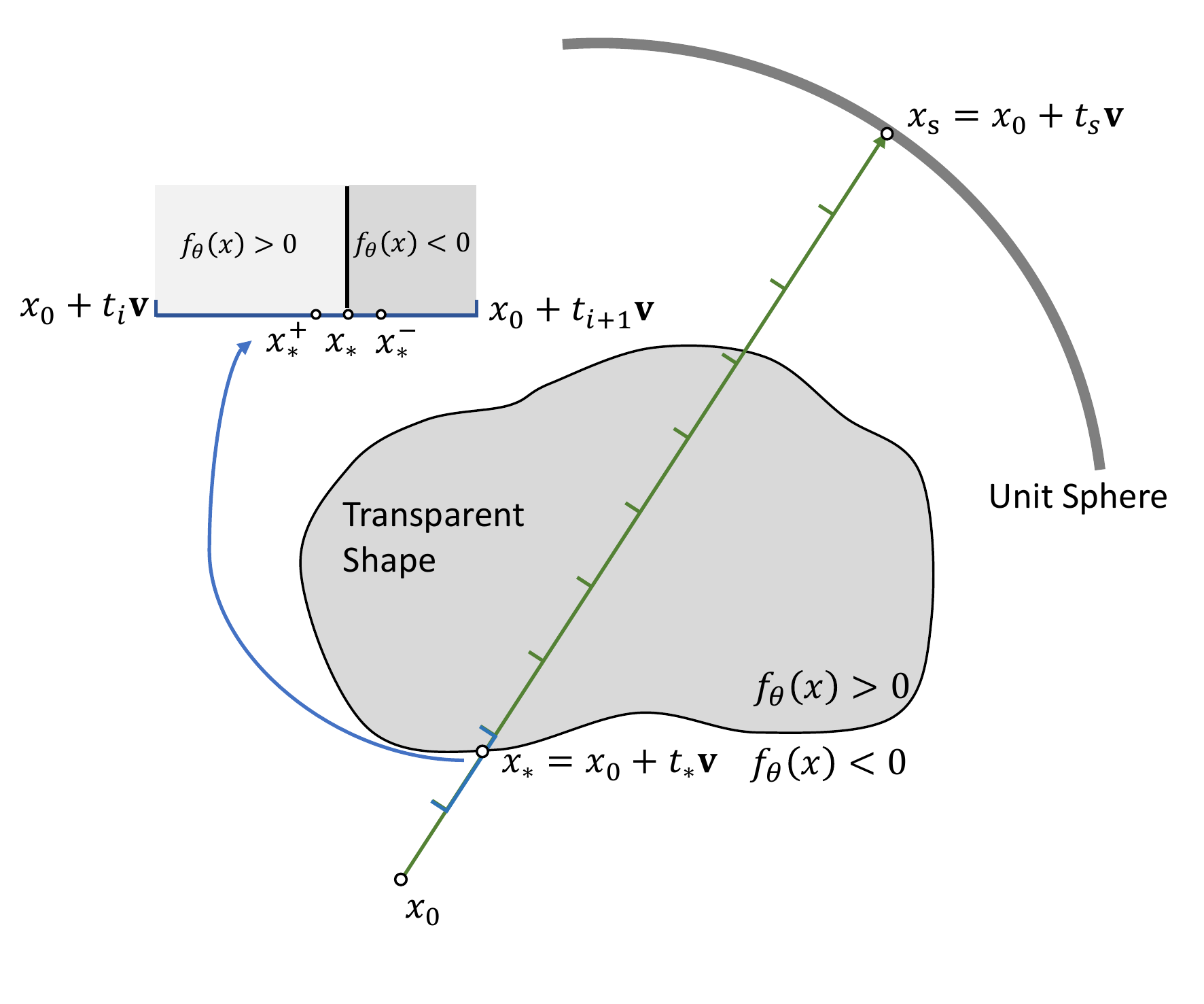} 
\caption{\textbf{Schematic diagram of the intersection calculation of the ray and SDF}} 
\label{fig:intersection} %
\end{figure}
As shown in Fig.\ref{fig:intersection}, we denote the ray as $x = x_0+t\mathbf{v}, t\geq0$, the intersection point of geometry as $x_*=x_0+t_*\mathbf{v}$ and the intersection point of unit sphere as  $x_s=x_0+t_s\mathbf{v}$, where $x_0$ is the start point of the ray and $\mathbf{v}$ is the unit direction vector. Similar to \cite{niemeyer2020differentiable}, we sample $100$ equal steps between $0$ and $t_s$, that is, $0<t_1<...<t_{99}<t_s$. Then we find the first $t_i$ and $t_{i+1}$ where $\sign{f_\theta(x_0+t_i\mathbf{v})}\neq\sign{f_\theta(x_0+t_{i+1}\mathbf{v})}$, the transition of signs of SDF values represents the ray crossing of the surface. The secant algorithm is used for approximation in the interval $(t_i,t_{i+1})$ and the nearest values to $t_*$ from both sides are recorded as $t_*^+$ and $t_*^-$, where $\sign{f_\theta(x_0+t_*^+\mathbf{v})}=\sign{f_\theta(x_0+t_{*}\mathbf{v})}$ and $\sign{f_\theta(x_0+t_*^-\mathbf{v})}\neq\sign{f_\theta(x_0+t_{*}\mathbf{v})}$. Both the $t_*^+$ and $t_*^-$ can be used as an approximation of $t_*$, the specific selection will depend on whether the next ray is refracted or reflected. The reason for calculating the two values of $t_*$ is that the sign of the start point will affect the correctness of the intersection result. For example, when $x_*^+=x_0+t_*^+\mathbf{v}$ is the start point of a refracted ray and its SDF value $f_\theta({x_*^+})>0$, the obtained initial interval must be $(t_0,t_1)$ since the refracted direction is toward the inside of the shape, resulting in a wrong intersection. The secant method with two interactions provides the basis for future multi-bounce($>2$) ray tracing in SDF, although only 2-bounce ray tracing is used in this paper.

\section{Dataset Details}
\subsection{Acquisition Details}
Our dataset consists of four objects' images from $34$ views, each of which contains four raw polarization images and three polarization parameters, intensity, degree of linear polarization(DoLP), and angle of linear polarization(AoLP), which are calculated from raw polarization images and will be described later. In addition, the mask and normal map of each view are also provided in our dataset. We used a 3D scanner to scan the powdered transparent objects to get the ground-truth shapes, then manually aligned them into each view to get the ground-truth normal maps.

Accurate camera poses for $34$ views are also provided, which are obtained by a precisely controlled robotic arm. When the origin of spherical coordinate setting to the center of the object, we uniformly sampled the views at the azimuths of $(0^\circ,340^\circ)$ with an interval of $20^\circ$ each time, and zenith angles of $50^\circ$ and $70^\circ$.
\subsection{Preprocessing}
We use a polarization camera to capture polarization images with four built in polarizer arrays of $0^\circ,45^\circ,90^\circ,135^\circ$. Therefore, four intensity images of $I_{0^\circ}, I_{45^\circ}, I_{90^\circ}, I_{135^\circ}$ can be obtained in a single shot, and the stokes vector can be written from the four intensities:
\begin{equation}
	\mathbf{s}= 
	\begin{bmatrix}
		s_0   \\
		s_1   \\
		s_2   \\
		s_3   \\		
	\end{bmatrix}=
		\begin{bmatrix}
			I_{0^\circ} + I_{90^\circ}   \\
			I_{0^\circ} - I_{90^\circ} \\
			I_{45^\circ} - I_{135^\circ} \\
			0   \\		
		\end{bmatrix}
\end{equation}
where $s_3$ represents the right circular polarzation component, which is zero here because the polarization camera can only capture the linear polarization state.

The degree of linear polarization(DoLP) $\rho$ and angle fo linear polarization(AoLP) $\psi$ are calculated from the following equations:
\begin{equation}
	\rho = \frac{\sqrt{s_1^2+s_2^2}}{s_0}
\end{equation}
\begin{equation}
	\psi = \frac{1}{2}\arctan(\frac{s_2}{s_1})
\end{equation}
The DoLP and AoLP maps are obtained by calculating the DoLP and AoLP values of each pixel and are used as the polarization part of our dataset.

\subsection{Illumination Sampling}
\begin{figure}[htbp]
\centering 
\includegraphics[width=\linewidth]{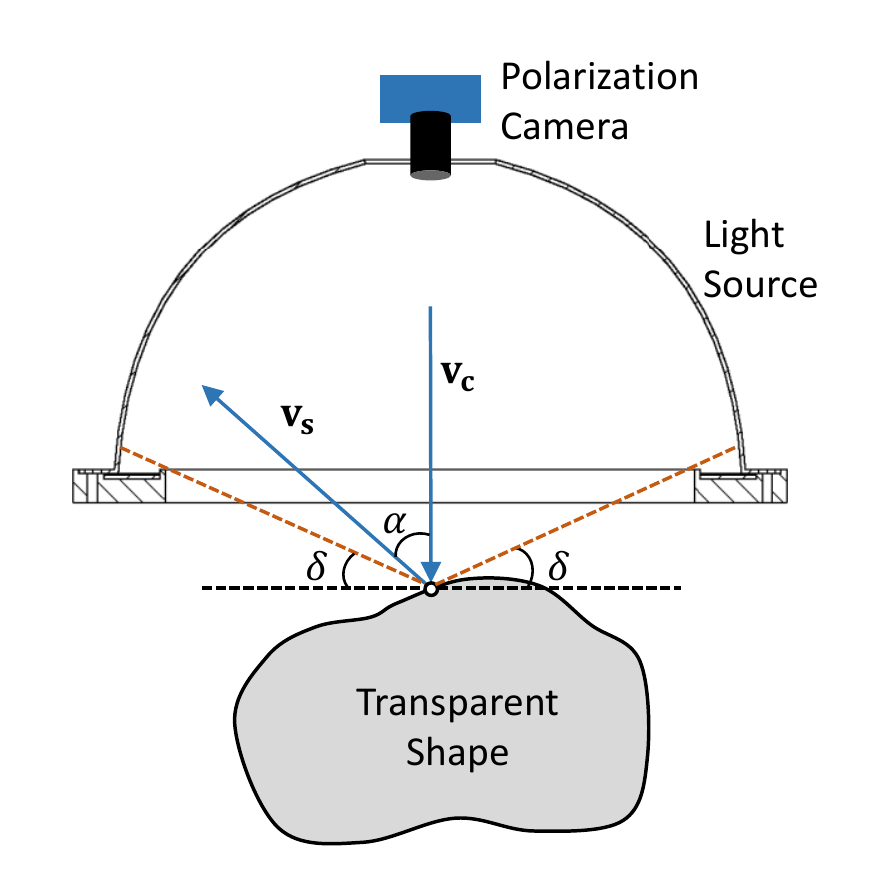} 
\caption{\textbf{Model for illumination sampling}} 
\label{fig:sampling} %
\end{figure}

A diffuse sphere is used as the light source in our dataset acquisition, which enhances the reflection component of the transparent surface. The illumination sampling is needed in our polar render and ray tracer. Hence, we model the illumination sampling as shown in Fig.\ref{fig:sampling}, where $\mathbf{v_c}$ is the view direction and $\mathbf{v_s}$ is the direction of the sampling ray(the ray that needs illumination sampling). 

The relative position between the light source and the polarization camera is fixed, so only the angle $\alpha$ between the direction of the sampled ray $v_s$ and the view direction $v_c$ can determine whether the sampled ray intersects the light source. Our illumination sampling function can be written as follows:
\begin{equation}
	\alpha = \arccos(\frac{-\mathbf{v_c}\cdot\mathbf{v_s}}{\Vert\mathbf{v_c}\Vert_2 \Vert\mathbf{v_s}\Vert_2})
\end{equation}
\begin{equation}
	I_{sample} = \begin{cases}
		1.0,& \text{ $ 0\leq \alpha < \frac{\pi}{2}-\delta$ } \\
		0.1,& \text{$otherwise$}
		\end{cases}
\end{equation}
where $I_{sample}$ represents the intensity of the sampling ray. We limit the angular range that can directly sample the light source to $[0,\frac{\pi}{2}-\delta)$, because the bottom of the light source can not fit the object surface and some rays within a range $\delta$ can not sample the light source directly. The default value of $\delta$ is $\frac{\pi}{18}$. We set the intensity of the light source to $1.0$, and the intensity of the other areas to $0.1$ since the environment has weak illumination from diffuse reflection and others.
\section{Transformations of Stokes Frames}
The value of the Stokes vector is related to the reference frame. To facilitate Mueller's calculation, we adopt the coordinate system and transformation similar to Mistuba2\cite{ravi2020accelerating}, which will be introduced in detail below. We denote the reference frame as $(x,y)$ omitting its $z$ representation, the $z$-axis is always along the ray direction.

\begin{figure}[htbp]
\centering 
\includegraphics[width=\linewidth]{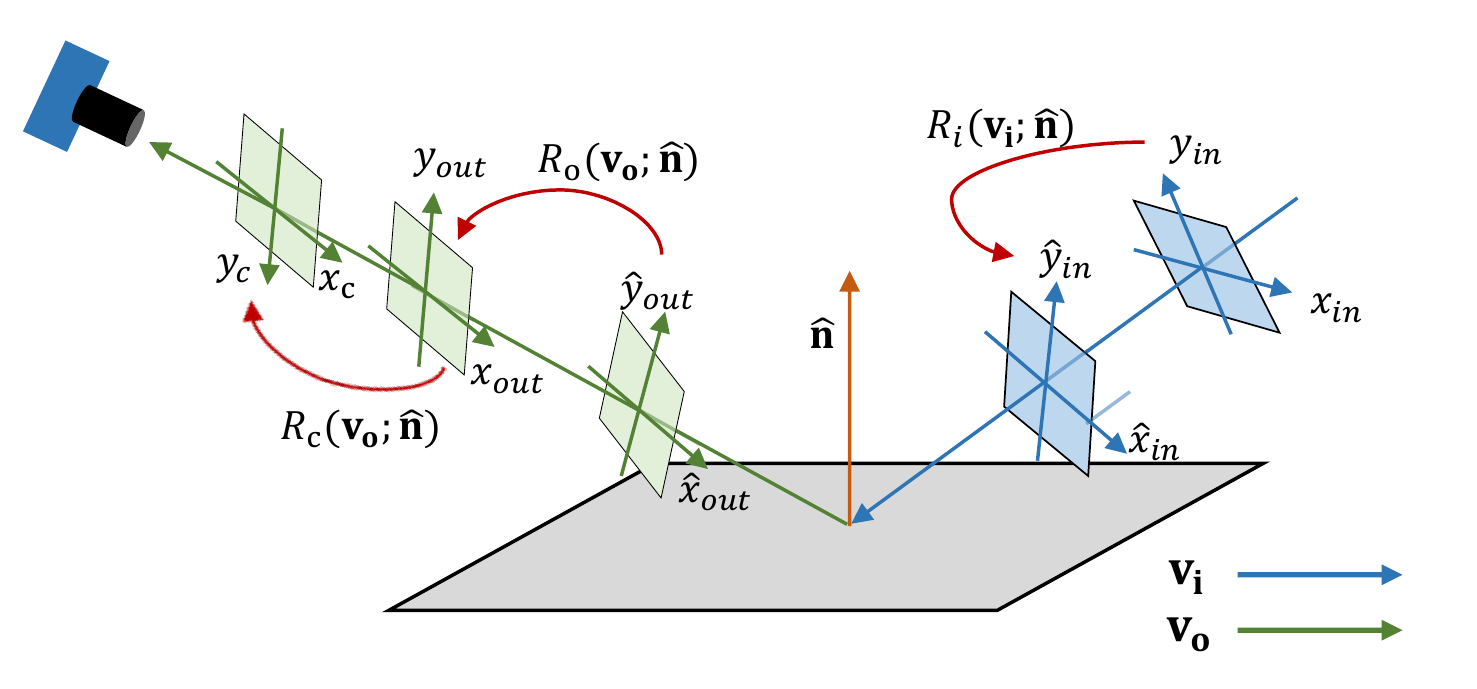} 
\caption{\textbf{Transfomations of incident and outgoing frames}} 
\label{fig:frames} %
\end{figure}
First, we define the $Rotator(\Delta \theta)$ similar to Mitsuba2, which represents the transformation matrix from frame $(x,y)$ to $(x',y')$  where $(x',y')$ is obtained by rotating $(x,y)$ around the $z-axis$ with angle of $\Delta\theta$:
\begin{equation}
	Rotator(\Delta \theta)= 
	\begin{bmatrix}
		1 & 0 & 0 & 0   \\
		0 & \cos(2\Delta\theta) & \sin(2\Delta\theta) & 0   \\
		0 & -\sin(2\Delta\theta) & \cos(2\Delta\theta) & 0   \\
		0 & 0 & 0 & 1   \\		
	\end{bmatrix}
\end{equation}
The transformation matrix is written in the form of $4\times4$ to match the size of the Mueller matrix.

The reference frames used in our polarimetric render are shown in Fig.\ref{fig:frames}. The frame $(x_{in},y_{in})$ is defined by the incident plane, where $x_{in}$ is perpendicular to the incident plane formed by $\mathbf{v_{in}}$ and $\mathbf{\hat{n}}$. Frames $(\hat{x}_{in},\hat{y}_{i})$ and $(\hat{x}_{out},\hat{y}_{out})$ are implicit orthogonal frames with $\mathbf{v_{i}}$ and $\mathbf{v_o}$ as $z$-axis, respectively. Using implicit orthographic frames enables us to extend our polarization renderer to multi-bounce polarimetric ray tracer easily in the future. $(x_{in},y_{in})$ and $(\hat{x}_{in},\hat{y}_{in})$ have the same $z$-axis and their transformation can be written by exploiting the definition of the $Rotator$:
\begin{equation}
	\Delta \theta_{in}= <x_{in},\hat{x}_{in}>=<\mathbf{\hat{n}
	\times{v}_i},Implicit(\mathbf{{v}_i)}>
\end{equation}
\begin{equation}
	R_i(\mathbf{v_i;\mathbf{\hat{n}}})= Rotator(\Delta \theta_{in})
\end{equation}
where $<\cdot>$ means the signed angle between the unit vectors. $Implicit(\cdot)$ represents the implicit orthographic frame calculation function, which takes the $z$-axis direction as input and outputs the $x$-axis direction vector.

Similarly, the transformation between $(x_{out},y_{out})$ and $(\hat{x}_{out},\hat{y}_{out})$ can be obtained:
\begin{equation}
	\Delta \theta_{out}= <x_{out},\hat{x}_{out}>=<Implicit(\mathbf{{v}_i)}
	,Implicit(\mathbf{{v}_o})>
\end{equation}
\begin{equation}
	R_o(\mathbf{v_o;\mathbf{\hat{n}}})= Rotator(\Delta \theta_{out})
\end{equation}

Finally, the Stokes vector in frame $(x_{out},y_{out})$ is needed to convert into the pixel coordinate frame $(x_c,y_c)$. The up direction$(-y_c)$ of the camera can be written as follows:
\begin{equation}
	\mathbf{up} =  P_c\begin{bmatrix}
			0 \\
			-1 \\
			0 \\
		\end{bmatrix}
\end{equation}
where $P_c$ is the extrinsic rotation matrix of the camera. With the representation in world coordinate frame of the camera's up direction, we can describe the transformation from $(x_{out},y_out)$ to $(x_c,y_c)$ using $Rotator$:
\begin{equation}
	\Delta \theta_{c}= <x_{out},x_c>=<Implicit(\mathbf{v_i)}
	,\mathbf{up}\times \mathbf{v_o}>
\end{equation}
\begin{equation}
	R_c(\mathbf{v_o})= Rotator(\Delta \theta_{c})
\end{equation}
Through the above frames transformations, the Stokes vector $\mathbf{s_0}$ emitted from the light source is observed in the camera as $\mathbf{s_c}$:
\begin{equation}
	\mathbf{s_c} =R_c(\mathbf{v_o})R_o(\mathbf{v_o;\mathbf{\hat{n}}})M_rR_i(\mathbf{v_i;\mathbf{\hat{n}}})s_0
\end{equation}
where $M_r$ is the Mueller matrix of reflection and its detailed expression is presented in the main text.

\end{document}